\documentclass[twocolumn, switch]{article}
\pdfoutput=1

\usepackage[text={174mm,258mm},%
papersize={210mm,297mm},%
columnsep=12pt,%
headsep=21pt,%
centering]{geometry}
\usepackage{ftnright}

\RequirePackage{fancyhdr}
\fancyheadoffset{0pt}
\RequirePackage{times}
\RequirePackage{caption}
\date{Published at~\citep{oelerichVDI}}

\usepackage[colorlinks,bookmarksopen,bookmarksnumbered,citecolor=red,urlcolor=red]{hyperref}
\usepackage{moreverb,url}
\usepackage{pgfplots}
\usepackage{pgfplotstable}
\usepgfplotslibrary{groupplots}
\usepackage{amsmath}
\usepackage{amsfonts}
\usepackage{bm}
\usepackage{import}
\usepackage{siunitx}
\usepackage{cleveref}
\usepackage{algorithm}
\usepackage{array}
\usepackage{algpseudocode}
\usepackage{textcomp}
\usepackage{stfloats}
\usepackage{verbatim}
\usepackage{graphicx}
\usepackage{caption}
\usepackage{subcaption}
\usepackage{cancel}

\usepackage[english]{babel}
\usepackage[utf8]{inputenc} % Deutsche Sonderzeichen/Umlaute gestatten
\usepackage[T1]{fontenc}
\usepackage{xcolor}
\usepackage{cancel}
\usepackage{enumitem}
\usepackage{moreverb,url}
\usepackage{paralist}

\usepackage{booktabs} % Erweiterte Tabellenumgebung

\usepackage{caption}
\usepackage[
    backend=biber,
    natbib=true,
    style=numeric-comp,
    sorting=none
]{biblatex}

\crefformat{figure}{Fig.~#2#1#3}
\Crefformat{figure}{Figure~#2#1#3}
\crefformat{equation}{(#2#1#3)}
\crefformat{section}{Section~#2#1#3}
\crefformat{table}{Table~#2#1#3}
\crefformat{algorithm}{Algorithm~#2#1#3}

\DeclareRobustCommand{\vec}[1]{
        \ifthenelse{\equal{#1}{\omega} \OR \equal{#1}{\varphi} \OR \equal{#1}{\alpha} \OR \equal{#1}{\beta} \OR \equal{#1}{\chi} \OR \equal{#1}{\delta} \OR \equal{#1}{\varepsilon} \OR \equal{#1}{\phi} \OR \equal{#1}{\epsilon} \OR \equal{#1}{\gamma} \OR \equal{#1}{\eta} \OR \equal{#1}{\iota} \OR \equal{#1}{\kappa} \OR \equal{#1}{\lambda} \OR \equal{#1}{\mu} \OR \equal{#1}{\nu} \OR \equal{#1}{\pi} \OR \equal{#1}{\theta} \OR \equal{#1}{\vartheta} \OR \equal{#1}{\rho} \OR \equal{#1}{\sigma} \OR \equal{#1}{\varsigma} \OR \equal{#1}{\tau} \OR \equal{#1}{\upsilon} \OR \equal{#1}{\xi} \OR \equal{#1}{\psi} \OR \equal{#1}{\zeta}}{
                \boldsymbol{#1}
        }{
                \mathbf{#1}
        }
}

\usetikzlibrary {3d}
\usetikzlibrary{decorations.markings}
\usepgfplotslibrary{fillbetween} 
\usetikzlibrary{external}
\usetikzlibrary {arrows.meta}
\pgfplotsset{every axis/.append style={
    font=\footnotesize,
    line join=round,
    legend style={/tikz/every even column/.append style={column sep=0.2cm}}
}}

\DeclareRobustCommand{\qs}[1]{\textsinglequote s}

\newcommand{\transpose}[1]{#1^\mathrm{T}}

\newcommand{\norm}[1]{\left\lVert#1\right\rVert}

\renewcommand{\dddot}[1]{%
{\mathop{\kern\z@#1}\limits^{\makebox[0pt][c]{\vbox to-1.4\ex@{\kern-\tw@\ex@
\hbox{\normalfont...}\vss}}}}}
\newcommand{\dddotsuper}[1]{%
{\mathop{\kern\z@#1}\limits^{\makebox[0pt][c]{\vbox to-1.6\ex@{\kern-\tw@\ex@
\hbox{\scriptsize...}\vss}}}}}
\newcommand{\dddotalt}[1]{%
{\mathop{\kern\z@#1}\limits^{\makebox[0pt][c]{\vbox to-2.2\ex@{\kern-\tw@\ex@
\hbox{\normalfont\scaleddot\kern-0.5pt\scaleddot\kern-0.5pt\scaleddot}\vss}}}}}
\makeatother

\usepackage{authblk}

\author[1]{Thies Oelerich}
\author[1]{Christian Hartl-Nesic}
\author[1, 2]{Andreas Kugi}
\affil[1]{Automation and Control Institute (ACIN), TU Wien, Vienna, Austria}
\affil[2]{AIT Austrian Institute of Technology GmbH, Vienna, Austria}
\begin{document}

\title{Model Predictive Trajectory Planning for Human-Robot Handovers}

%\affiliation{\affilnum{1}Automation and Control Institute (ACIN), TU
%Wien, Vienna, Austria,
%\\
%\affilnum{2}AIT Austrian Institute of Technology GmbH, Vienna, Austria}

\twocolumn[ % Method A for two-column formatting
  \begin{@twocolumnfalse} % Method A for two-column formatting
  
\maketitle
\begin{abstract}
    This work develops a novel trajectory planner for human-robot handovers. The
    handover requirements can naturally be handled by a path-following-based
    model predictive controller, where the path progress serves as a progress
    measure of the handover. Moreover, the deviations from the path are used to
    follow human motion by adapting the path deviation bounds with a handover
    location prediction. A Gaussian process regression model, which is trained
    on known handover trajectories, is employed for this prediction. Experiments
    with a collaborative 7-DoF robotic manipulator show the effectiveness and
    versatility of the proposed approach.
\end{abstract}
\vspace{0.35cm}

  \end{@twocolumnfalse} % Method A for two-column formatting
] % Method A for two-column formatting

%\keywords{Model Predictive Control, Trajectory Planning, Robotic Manipulator,
%Path Following, Online Replanning}

\maketitle

\newcommand{\goalconv}{(\textit{Convergence})}
\newcommand{\goalsync}{(\textit{Synchronization})}
\newcommand{\goalpredict}{(\textit{Predictability})}
\newcommand{\goalsafe}{(\textit{Safety})}
\newcommand{\goalconvnb}{\textit{Convergence}}
\newcommand{\goalsyncnb}{\textit{Synchronization}}
\newcommand{\goalpredictnb}{\textit{Predictability}}
\newcommand{\goalsafenb}{\textit{Safety}}

\section{Introduction}

As robots become more capable at a high pace, they are expected to solve
increasingly complex problems. Dynamically changing environments are challenging
for robots to
navigate and act in. A demanding and vital field in this context is
human-robot interaction. As human intentions are hard to model, robots
need to be able to quickly and reliably react to human motions while respecting
dynamic and kinematic constraints. Human-robot handovers are a basic yet complex
case of human-robot interactions.

This work uses a novel path-following-based trajectory planner for a robotic
manipulator to realize natural human-robot handovers. A motion planning
problem is solved online to enable the robot to react to the human behavior and
the predictions of the human behavior.

% MPC 
Motion planning in robotics is a well-studied problem where many solution
methods exist. Popular solutions are based on
optimization~\citep{schoelsCIAOMPCbasedSafe2020,
schulmanMotionPlanningSequential2014},
sampling~\citep{jankowskiVPSTOViapointbasedStochastic2023,
perssonSamplingbasedAlgorithmRobot2014, elbanhawiSamplingBasedRobotMotion2014},
or learning~\citep{osaMotionPlanningLearning2022,
carvalhoMotionPlanningDiffusion2023}. This work focuses on optimization-based
planning, precisely  model predictive control (MPC), which is a popular
optimization-based framework to plan motions and respect
constraints~\citep{schoelsCIAOMPCbasedSafe2020, ardakaniRealtimeTrajectoryGeneration2015, yangModelPredictiveControl2022}.
It uses a
receding time horizon to plan locally optimal trajectories based on a cost
function at each time step. 
Path-following control is the first choice when a task is specified as a
geometric path to be followed, as in~\citep{astudilloVaryingRadiusTunnelFollowingNMPC2022,
vanduijkerenPathfollowingNMPCSeriallink2016}. Such a formulation allows
explicit control over the time evolution of the robot's motion, such as
controlling deviations from the path based on specific directions, as done
in~\citep{hartl-nesicSurfacebasedPathFollowing2021}.
This paper proposes a path-following formulation to implement handover
requirements naturally. Human-robot handovers can be interpreted as a
path-following problem with a continuously changing and uncertain path. As the
latter complicates the path-following task, existing solutions using MPC for
handovers consider the handover point as the final goal but do not have a
reference path. In~\citep{yangModelPredictiveControl2022}, a grasping
configuration is first computed, which is then reached by a suitable cost
function in the MPC. A similar cost function is also used
in~\citep{kshirsagarTimingSpecifiedControllersFeedback2022} with
a desired reach time. A more advanced objective, utilizing a tracking cost,
ergonomics, safety, and visibility, is employed
in~\citep{corsiniNonlinearModelPredictive2022}. This work also applies a
constant-velocity model to predict the human motion.

% Prediction of Human motion / handover point
Motion planning in a human-robot handover scenario significantly benefits from
predictions of the human motions. Thus, the path-following MPC formulation used
in this work must react online to the human movements. Model predictive
controller systematically incorporate such a behavior. A model for human
motion is considered in
addition to a model of the handover location. The handover location consists of
a position and an orientation, which is the desired
final pose of the robot\textquotesingle s end-effector. It is unknown prior to
the handover and may change during the handover.
Previous works commonly uses Gaussian process regression (GPR)
to model unknown dynamics, as in~\citep{hewingCautiousModelPredictive2020,
bethgeModelPredictiveControl2023}, which provides a
predictive distribution instead of a most likely next state. 
GPR is used in~\citep{liProvablySafeEfficient2021,
wuOnlineMotionPrediction2019} for human-robot interaction to predict human motions during the interaction.
Utilizing dynamic movement primitives,~\citep{widmannHumanMotionPrediction2018}
manages to predict human handover trajectories. Several deep-learning methods for
motion predictions are reviewed in~\citep{lyu3DHumanMotion2022}.

% Current work and contributions
In this work, a path-following-based MPC is combined with a GPR model
to predict the handover location and its uncertainty
measure. The path-following formulation is exploited to achieve successful
handovers. 
The decomposition of the motion in the path direction and orthogonal to it
allows us to interpret the motion along the path as the progress of the handover
task and to use the directions orthogonal to the path for online adaptions and
to synchronize the motion between the human and the robot.

The main contributions of this paper are as follows:
\begin{compactitem}
    \item the application of path-following MPC to human-robot
        handovers where the robot can move forward and backwards along the path
        based on the human motion,
    \item systematically taking into account the uncertainty of the handover
        location, and
    \item the adaption of the path error bounds during the handover to guide
        the robot towards the handover point and keep the human safe. 
\end{compactitem}

\section{Successful Handovers}
\label{sec:requirements_handover}

Human-robot handovers are complex interactions where the two actors, robot and
human, have to be able to react to each other and move in a partially
predictable way. The most essential requirements for a successful handover,
inspired by~\citep{yangModelPredictiveControl2022,
zacharakiSafetyBoundsHuman2020, strabalaSeamlessHumanRobotHandovers2013}, are:

\begin{compactitem}
    \item \goalconvnb: The handover motions of both actors end at a common pose
        called the handover location.
    \item \goalsyncnb: The time for the handover
        location is synchronized such that both actors reach it simultaneously.
    \item \goalpredictnb: The movements of each actor is predictable to let the
        other actor infer their intentions.
    \item \goalsafenb: Any collisions between the two actors are avoided.
\end{compactitem}

A path-following-based formulation naturally fulfills these
requirements. This formulation is further detailed in \cref{sec:mpc_extensions}
based on the controller outlined in \cref{sec:bound_mpc_formulation}.

\section{BoundMPC}
\label{sec:bound_mpc_formulation}

\begin{figure}
    \scriptsize\centering
    \def\svgwidth{\linewidth}
    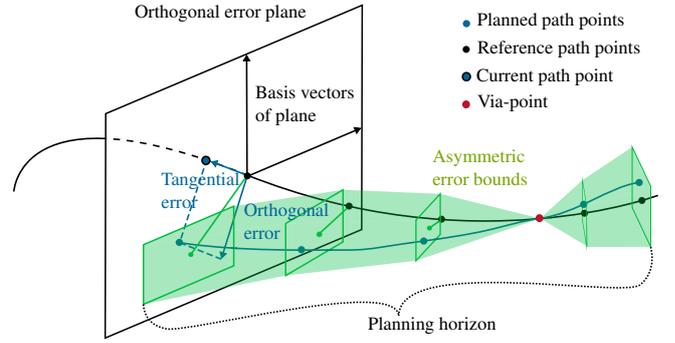
    \caption{Schematic working principle of BoundMPC~\citep{oelerichBoundMPCCartesianTrajectory2024}.}
    \label{fig:mpc_scheme}
\end{figure}

In this section, a short overview of the path-following MPC is given.  The
formulation used in this work is inspired by a framework developed by the
authors in a previous work called
BoundMPC~\citep{oelerichBoundMPCCartesianTrajectory2024}. A schematic is shown in
\textbf{\cref{fig:mpc_scheme}}. In this formulation, the desired robot motion is
given by
position and orientation reference paths $\vec{\pi}_{\mathrm{p}}(\phi)$ and
$\vec{\pi}_{\mathrm{o}}(\phi)$ with the path parameter $\phi$, which must be followed by the
robot\textquotesingle s end-effector with position $\vec{p}_{\mathrm{r}}(t)$ and
rotation matrix $\vec{R}_{\mathrm{r}}(t)$ in
Cartesian space. Bounds limit the deviation of the planned trajectory from the
reference path. Thus, the formulation makes the robot\textquotesingle s
end-effector move through a bounded volume in Cartesian space.  As a special
case, piecewise linear reference paths are used to connect the via-points. 

The core of the formulation is based on a decomposition of the position path
error $\vec{e}_{\mathrm{p, r}} = \vec{p}_{\mathrm{r}}(t) - \vec{\pi}_{\mathrm{p}}(\phi)$ and
orientation path error $\vec{e}_{\mathrm{o, r}}$ into tangential and orthogonal
directions. The position path error $\vec{e}_{\mathrm{p, r}}$
is decomposed into the direction $\vec{\pi}'_{\mathrm{p}}$
tangential to the
path and the two basic directions $\vec{b}_{\mathrm{p}, 1}$ and $\vec{b}_{\mathrm{p},
2}$ orthogonal to the path, where $(\cdot)'$ denotes the partial derivative
w.r.t. the path parameter $\phi$. This results in the representation
\begin{equation}
    \transpose{\vec{e}_{\mathrm{p, r}, \phi}} = \begin{bmatrix}
        e_{\mathrm{p, r}, 1}^{\bot} &
        e_{\mathrm{p, r}, 2}^{\bot} &
        e_{\mathrm{p, r}}^{||}
    \end{bmatrix}
\end{equation}
in the path system. BoundMPC minimizes the tangential error $e_{\mathrm{p, r}}^{||}$
and bounds the orthogonal errors $e_{\mathrm{p, r}, 1}^{\bot}$ and
$e_{\mathrm{p, r}, 2}^{\bot}$. Thus, the robot can
exploit the orthogonal errors to optimize its end-effector trajectory while
following the path. 
The orientation path error $\vec{e}_{\mathrm{o, r}} =
\mathrm{Log}(\vec{R}_{\mathrm{r}}(t)
\transpose{\mathrm{Exp}(\vec{\pi}_{\mathrm{o}}(\phi))})$ is obtained using matrix
exponential and logarithm functions, it is defined in the Lie space for
orientations; see~\citep{solaMicroLieTheory2021} for more details on this concept.
Decomposition is performed using the direction
$\vec{b}_{\omega_\mathrm{r}}$ along the path and the basic directions
$\vec{b}_{\mathrm{o}, 1}$ and $\vec{b}_{\mathrm{o}, 2}$ orthogonal to the path.
Hence, the orientation error in the
path system reads as
\begin{equation}
    \transpose{\vec{e}_{\mathrm{o, r}, \phi}} = \begin{bmatrix}
        e_{\mathrm{o, r}, 1}^{\bot} &
        e_{\mathrm{o, r}, 2}^{\bot} &
        e_{\mathrm{o, r}}^{||}
    \end{bmatrix} \text{ ,}
\end{equation}
where the individual components $e_{\mathrm{o, r}, 2}^{\bot}$,
$e_{\mathrm{o, r}}^{||}$, and $e_{\mathrm{o, r}, 1}^{\bot}$ are the roll, pitch,
and yaw angles, respectively, in the coordinate system defined by the
orientation direction vectors.

For more information, the reader is referred to~\citep{oelerichBoundMPCCartesianTrajectory2024}

\section{Extensions for Human-Robot Handovers}
\label{sec:mpc_extensions}

The previous work~\citep{oelerichBoundMPCCartesianTrajectory2024} is extended by adapting the MPC such
that a handover between a human and a robotic manipulator is possible according to
the requirements formulated in \cref{sec:requirements_handover}. The prediction
of the handover
location is detailed in \cref{ssec:handover_prediction}, and convergence of the
robot\textquotesingle s motion is ensured by adapting the error bounds
described in
\cref{ssec:error_bounds} and adding a terminal cost described in
\cref{ssec:terminal_cost} \goalconv. The error bounds constrain the
robot\textquotesingle s motion, allowing only limited motion orthogonal to the
path \goalpredict.
Furthermore, combining the desired path state with error bounds avoids
collisions between humans and robots, as discussed in
\cref{ssec:error_bounds} \goalsafe. Lastly, the path progress synchronizes the
humans and robots motions as described in \cref{ssec:desired_path_state}
\goalsync. 

\begin{figure}
    \centering
    \def\svgwidth{\linewidth}
    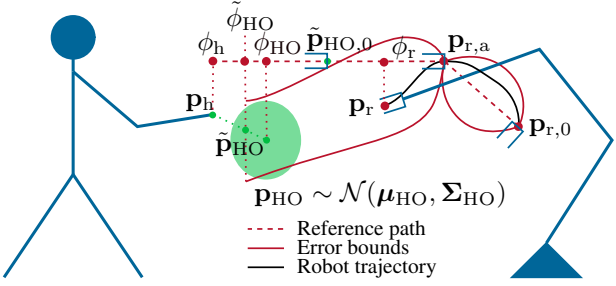
    \caption{Visualization of the considered human-robot handover situation. The
        predicted handover location distribution $\vec{p}_{\mathrm{HO}}$ which determines the error
        bounds is visualized by the green circle. As the
        human is already close the handover location, the adapted handover
        location $\tilde{\vec{p}}_{\mathrm{HO}}$ is a linear interpolation
        between the current human position $\vec{p}_{\mathrm{h}}$ and the
        predicted position $\vec{\mu}_{\mathrm{HO}}$. The reference orientations
        at
        $\tilde{\vec{p}}_{\mathrm{HO, 0}}$, $\vec{p}_{\mathrm{r, a}}$ and
    $\vec{p}_{\mathrm{r, 0}}$ are indicated by a gripper symbol.}
    \label{fig:handover_visualization}
\end{figure}

The handover scenario is depicted in \textbf{\cref{fig:handover_visualization}} with the
predicted handover location distribution $\vec{p}_{\mathrm{HO}} \sim
\mathcal{N}(\vec{\mu}_\mathrm{HO},\vec{\Sigma}_\mathrm{HO})$ and the current
position of the human hand $\vec{p}_{\mathrm{h}}$ and the robot\textquotesingle
s end-effector $\vec{p}_{\mathrm{r}}$. The adapted handover location used by the
MPC $\tilde{\vec{p}}_{\mathrm{HO}}$ is given by an interpolation between
$\vec{\mu}_{\mathrm{HO}}$ and $\vec{p}_{\mathrm{h}}$. As the human hand
approaches the handover location, a prediction is no longer needed, and thus
$\tilde{\vec{p}}_{\mathrm{HO}} \to \vec{p}_{\mathrm{h}}$. On the contrary, when
$\vec{p}_{\mathrm{h}}$ is far from $\vec{p}_{\mathrm{HO}}$, the handover
location can only be inferred from the prediction model, meaning
$\tilde{\vec{p}}_{\mathrm{HO}} \to \vec{\mu}_{\mathrm{HO}}$. This is further
detailed in \cref{sssec:prediction_adaptions}.

\subsection{Optimal control problem}
\label{ssec:mpc_formulation}

In this section, the optimal control problem from the previous work~\citep{oelerichBoundMPCCartesianTrajectory2024} is
adapted for human-robot handovers, which reads as
\begin{subequations}
\label{eq:mpc_problem}
    {
    \begin{align}
    \label{eq:mpc_objective} \min_{\substack{\vec{u}_{1}, \ldots, \vec{u}_{N}, \\ \vec{x}_{1}, \ldots,
    \vec{x}_{N}, \\ \vec{\xi}_{1}, \ldots, \vec{\xi}_{N}, \\ v_{1},
\ldots, v_{N}}}
    & J_{\mathrm{T}}(\vec{x}_{N}, \vec{\xi}_{N}, \vec{u}_{N},
v_{N}) + \sum_{i=1}^{N}  l(\vec{x}_{i}, \vec{\xi}_{i}, \vec{u}_{i}, v_{i}) \\
        \text{s.t.\quad}
                            \begin{split}
                            \label{eq:mpc_state} \vec{x}_{i+1} = 
                            \vec{\Phi} \vec{x}_{i} + \vec{\Gamma}_{0}
                            \vec{u}_{i} +& \vec{\Gamma}_{1} \vec{u}_{i+1}
                            \text{,~}
                            i = 0, \ldots, N-1 \\
                            \end{split} \\
                            \begin{split}
                                \label{eq:mpc_path_state} \vec{\xi}_{i+1}
                                =\vec{\Phi}_{\xi} \vec{\xi}_{i} + \vec{\Gamma}_{0,
                                \xi} v_{i} &+ \vec{\Gamma}_{1, \xi} v_{i+1}\text{,}\\
                                &i = 0, \ldots, N-1 \\
                            \end{split} \\
                            %& \label{eq:mpc_ref_p} \vec{\pi}_{\mathrm{p}, i+1} =
                            %\vec{\pi}_\mathrm{p}(\phi_{i+1})\\
                            %& \label{eq:mpc_ref_o} \vec{\pi}_{\mathrm{o}, i+1} =
                            %\vec{\pi}_\mathrm{o, 0} +
                            %\vec{J}_\mathrm{l}^{-1}(\vec{\pi}_\mathrm{o, 0})
                            %\Delta \vec{\Omega}_{i+1}(\phi_{i+1})\\
                             %&\label{eq:mpc_path_forward} \cancel{\dot{\phi}_{i+1} \geq 0} \\
                             & \label{eq:mpc_init_x} \vec{x}_{0} = \vec{x}_{\mathrm{init}} \\
                            & \vec{u}_{0} = \vec{u}_{\mathrm{init}} \\
                            & \vec{\xi}_{0} = \vec{\xi}_{\mathrm{init}} \\
                            & \label{eq:mpc_init_v} v_{0} = v_{\mathrm{init}} \\
                            & \label{eq:mpc_bound_x} \underline{\vec{x}} \leq \vec{x}_{i+1} \leq \overline{\vec{x}}\\
                            & \label{eq:mpc_bound_u} \underline{\vec{u}} \leq \vec{u}_{i+1} \leq \overline{\vec{u}} \\
                            & \label{eq:mpc_path_vel}\dot{\phi}_{\mathrm{d}} = \dot{\phi}_{\mathrm{d, max}} \tanh\left(s (d_{\mathrm{r, HO}}
                            - d_{\mathrm{h, HO}} + b)\right) \\
                            \begin{split}
                            \label{eq:mpc_err_pos} \Psi_{\mathrm{p}, m,
                                \mathrm{lower}}(\phi_i) \leq e_{\mathrm{p, r},
                        m}^{\bot} \leq \Psi_{\mathrm{p}, m,
                    \mathrm{upper}}(\phi_i)\text{,~} \\ m = 1, 2\\
                            \end{split} \\
                            \begin{split}
                            \label{eq:mpc_err_rot} \Psi_{\mathrm{o}, m,
                                \mathrm{lower}}(\phi_i) \leq e_{\mathrm{o, r},
                        m}^{\bot} \leq \Psi_{\mathrm{o}, m,
                    \mathrm{upper}}(\phi_i) \text{,~}\\m = 1, 2\text{.}
                            \end{split}
                 \end{align}}
\end{subequations}
The robot joint
and path dynamics are the discrete-time linear
systems~\cref{eq:mpc_state} and~\cref{eq:mpc_path_state}, respectively. The states and inputs
for these systems have the
initial conditions~\cref{eq:mpc_init_x}-\cref{eq:mpc_init_v} and
bounds~\cref{eq:mpc_bound_x} and~\cref{eq:mpc_bound_u}. 

The extensions to the previous work are the motion
synchronization~\cref{eq:mpc_path_vel} explained in
\cref{ssec:desired_path_state}, the bounds~\cref{eq:mpc_err_pos}
and~\cref{eq:mpc_err_rot} explained in \cref{ssec:error_bounds}, and the removal
of the constraint $\dot{\phi}_{i+1} \geq 0$, which allows the robot to move
backwards along the path, i.e., $\dot{\phi}_{i+1} \leq 0$ is permitted .
Furthermore, the cost function~\cref{eq:mpc_objective} is extended by the
terminal cost $J_{\mathrm{T}}(\vec{x}_{N}, \vec{\xi}_{N}, \vec{u}_{N}, v_{N})$.
The stage cost $l(\vec{x}_{i}, \vec{\xi}_{i}, \vec{u}_{i}, v_{i})$ is still used
to minimize the tangential path error and reach the desired path state
$\vec{\xi}_{\mathrm{d}}$. For more information, see~\citep{oelerichBoundMPCCartesianTrajectory2024}.

\subsection{Reference paths}
\label{ssec:reference_paths}

The reference paths for the robot\textquotesingle s end-effector
$\vec{\pi}_{\mathrm{p}}(\phi)$ and $\vec{\pi}_{\mathrm{o}}(\phi)$ are determined
at the beginning of the handover and are kept constant during the handover. This
is based on the assumption that the handover location distribution is
sufficiently accurate
and that the actual handover location will not deviate significantly. The robot is
assumed to start at a specific position $\vec{p}_{\mathrm{r}, 0}$ from which it will
move through an approach
position $\vec{p}_{\mathrm{r, a}}$ towards the handover location
$\tilde{\vec{p}}_{\mathrm{HO}}$, see \cref{fig:handover_visualization}. The
linear path
from $\vec{p}_{\mathrm{r, 0}}$ to the initially estimated handover location
$\tilde{\vec{p}}_{\mathrm{HO}, 0}$ is used as the reference path
$\vec{\pi}_{\mathrm{p}}(\phi)$.
The actual
handover location will generally deviate from the path, which is accounted for by
adapting the error bounds in \cref{ssec:error_bounds} and the terminal
costs in \cref{ssec:terminal_cost}. This approach allows the use of the path
parameter $\phi$ as the progress of the handover. Thus, the path velocity
$\dot{\phi}$ serves to synchronize the robot and human motion.
The orientation reference path $\vec{\pi}_{\mathrm{o}}(\phi)$ is also determined
from the initially estimated handover location $\tilde{\vec{p}}_{\mathrm{HO}, 0}$. The reference
path $\vec{\pi}_{\mathrm{o}}(\phi)$ is
chosen such that the
gripper is oriented towards the human hand at $\tilde{\vec{p}}_{\mathrm{HO},
0}$,
as indicated by the gripper symbols in \cref{fig:handover_visualization}.

%\begin{figure}
%    \centering
%    \def\svgwidth{\linewidth}
%    \import{inkscape}{reference_paths.pdf_tex}
%    \caption{Reference paths during the human-robot handover. Dots represent the
%        via-points which are connected using linear segments. The orientation
%        reference is shown as the blue grippers on the via-points. The green
%        arrow indicates that the final reference path point
%        $\tilde{\vec{p}}_{\mathrm{HO}}$ is determined based on the human
%    position $\vec{p}_{\mathrm{h}}$ using the prediction model.}
%    \label{fig:reference_paths}
%\end{figure}

\subsection{Handover Location Prediction}
\label{ssec:handover_prediction}

As stated in \cref{sec:requirements_handover}, both actors must
move to a mutual handover location \goalconv. However, this location is unknown
to the robot, and thus, a prediction model is needed to estimate it. In this
section, a Gaussian process model is proposed to predict the handover location
$\vec{p}_{\mathrm{HO}}$ in \cref{sssec:gaussian_process}. As the human hand
approaches the actual handover location, this prediction gradually switches
towards the actual position of the human hand $\vec{p}_{\mathrm{h}}$, as detailed
in \cref{sssec:prediction_adaptions}.

\subsubsection{Gaussian Process Model}
\label{sssec:gaussian_process}

The purpose of this model is to predict the distribution of the position of the Cartesian
handover location $\vec{p}_\mathrm{HO} \in \mathbb{R}^{3}$ from the
measured current hand position $\vec{p}_{\mathrm{h}}$. This work uses a Gaussian
process regression model for this task. Thus, the distribution of
$\vec{p}_{\mathrm{HO}}$ is given by the Gaussian distribution 
\begin{equation}
    \vec{p}_\mathrm{HO} \sim \mathcal{N}(\vec{\mu}_\mathrm{HO},
    \vec{\Sigma}_\mathrm{HO})\text{~,}
\end{equation}
with mean $\vec{\mu}_{\mathrm{HO}}$ and covariance matrix
$\vec{\Sigma}_{\mathrm{HO}}$.
By assuming that the components of $\vec{p}_{\mathrm{HO}}$ are independent, it
is possible to train separate prediction models for each spatial component.
The training data for each of the three models $j \in [\mathrm{x, y, z}]$ consists of $N$
known handover trajectories with $n_{k}$ data points in the $k$-th trajectory
resulting in a total of $n = \sum_{k = 1}^{N} n_{k}$ data points.
All data points are summarized in the input matrix $\vec{X}_{\mathrm{train}}
\in \mathbb{R}^{n \times
6}$ and the output vector $\vec{y}_{\mathrm{train}} \in \mathbb{R}^{n}$, where
each row is given by
\begin{equation}
    (\transpose{\vec{x}_{\mathrm{train}, i}}, y_{\mathrm{train}, i}) = \left(\begin{bmatrix}
        \vec{p}_{\mathrm{h, train}, i}\\ 
        \vec{v}_{\mathrm{h, train}, i}
    \end{bmatrix}^{\mathrm{T}}, 
p_{\mathrm{HO, train}, j, i}\right) \text{~,}
\end{equation}
with $\vec{x}_{\mathrm{train}, i}$ containing the  human position $\vec{p}_{\mathrm{h, train}, i} \in \mathbb{R}^{3}$
and velocity $\vec{v}_{\mathrm{h, train}, i} \in \mathbb{R}^{3}$ , $i = 1,
\ldots, n$ 
and the corresponding output handover location
\begin{equation}
    \transpose{\vec{p}_{\mathrm{HO, train}, i}} = \begin{bmatrix}
        p_{\mathrm{HO, train, x}, i},
        p_{\mathrm{HO, train, y}, i},
        p_{\mathrm{HO, train, z}, i} \\ 
    \end{bmatrix} \text{ .}
\end{equation}
Given a scalar kernel function
$k(\vec{x}_{\mathrm{train}, i}, \vec{x}_{\mathrm{train}, j})$, the matrix
$\vec{K}(\vec{X}_{\mathrm{train}}, \vec{X}_{\mathrm{train}}) \in \mathbb{R}^{n \times
n}$ is the result of
applying this kernel function to all pairs of training input points
$\vec{x}_{\mathrm{train}, i}$ contained in $\vec{X}_{\mathrm{train}}$. In this
work, the squared exponential kernel function is used. 
A test input point $\vec{x}_{\mathrm{test}}$ then leads to the predicted output
mean and covariance~\citep{rasmussenGaussianProcessesMachine2006}
\begin{subequations}
    \label{eq:prediction_model}
    \begin{align}
        \mu_{\mathrm{HO}, j} &= \transpose{\vec{k}}_{\mathrm{test}} (\vec{K} +
        \sigma^2_{\mathrm{n}} \vec{I})^{-1} \vec{y}_{\mathrm{train}} \\ 
        \Sigma_{\mathrm{HO}, j} &= k(\vec{x}_{\mathrm{test}},
        \vec{x}_{\mathrm{test}}) - \transpose{\vec{k}}_{\mathrm{test}} (\vec{K} +
        \sigma_{\mathrm{n}}^2 \vec{I})^{-1} \vec{k}_{\mathrm{test}}
    \end{align}
\end{subequations}
with $\vec{k}_{\mathrm{test}}$ being the vector of kernel function values
between the test point $\vec{x}_{\mathrm{test}}$ and all training points
$\vec{x}_{\mathrm{train}, i}$, and $\sigma^2_{\mathrm{n}}$ is the variance of
the assumed zero mean measurement noise of $\vec{y}_{\mathrm{train}}$. The
covariance values $\Sigma_{\mathrm{HO}, j}$ determine the diagonal of
$\vec{\Sigma}_{\mathrm{HO}}$. All other values of $\vec{\Sigma}_{\mathrm{HO}}$
are zero due to the assumed independence. For more information on Gaussian
process regression, the reader is referred
to~\citep{rasmussenGaussianProcessesMachine2006}.

\subsubsection{Projection of the Handover Prediction}
\label{sssec:projection_handover_location}
The prediction of the handover location distribution $\vec{p}_{\mathrm{HO}}$ is
done in the global coordinate system. In order to use it within the path-following
MPC, the handover location distribution is projected onto the path
$\vec{\pi}_{\mathrm{p}}$ using the basis vectors $\vec{b}_{\mathrm{p}, 1}$ and
$\vec{b}_{\mathrm{p}, 2}$ of the
path segment in which the handover takes place. Its mean
is represented as 
\begin{equation}
    \label{eq:path_representation}
    \transpose{\vec{\mu}_{\mathrm{HO, \phi}}} = \begin{bmatrix}
        e^{\bot}_{\mathrm{p, HO}, 1}&
        e^{\bot}_{\mathrm{p, HO}, 2}&
        \phi_{\mathrm{HO}}
    \end{bmatrix}
\end{equation}
with
\begin{subequations}
    \label{eq:path_projection}
    \begin{align}
        \phi_{\mathrm{HO}} &= \phi_0 + (\transpose{(\vec{\mu}_{\mathrm{HO}} -
        \vec{\pi}_{\mathrm{p}}(\phi_0))}
        \vec{\pi}_{\mathrm{p}}'(\phi_{\mathrm{HO}})) \\
        \begin{bmatrix}
            e^{\bot}_{\mathrm{p, HO}, 1} \\
            e^{\bot}_{\mathrm{p, HO}, 2}
        \end{bmatrix} &= 
        \begin{bmatrix}
            \transpose{\vec{b}_{\mathrm{p}, 1}} \\
            \transpose{\vec{b}_{\mathrm{p}, 2}}
        \end{bmatrix} 
        (\vec{\mu}_{\mathrm{HO}} - \vec{\pi}(\phi_{\mathrm{HO}})) \text{ ,}
    \end{align}
\end{subequations}
where $e^{\bot}_{\mathrm{p, HO}, 1}$ and $e^{\bot}_{\mathrm{p, HO}, 2}$ are the
orthogonal deviations from the path, $\phi_{\mathrm{HO}}$ is the path parameter
at which the handover location takes place, $\vec{\pi}(\phi_{\mathrm{HO}})$ is
the point on the reference path closest to $\vec{\mu}_{\mathrm{HO}}$, and
$\phi_{0}$ is the path parameter at $\vec{p}_{\mathrm{r, a}}$.

In addition to the mean $\vec{\mu}_{\mathrm{HO}}$, the prediction model from
\cref{ssec:handover_prediction} also provides the uncertainty
$\vec{\Sigma}_{\mathrm{HO}}$ of the handover location in
the global coordinate system. The basis vectors $\vec{b}_{\mathrm{p}, 1}$ and
$\vec{b}_{\mathrm{p}, 2}$ are used to obtain 
\begin{equation}
    \vec{\Sigma}_{\mathrm{HO}, \phi} = 
    \begin{bmatrix}
        \transpose{\vec{b}_{\mathrm{p}, 1}} \\
        \transpose{\vec{b}_{\mathrm{p}, 2}}
    \end{bmatrix} 
    \vec{\Sigma}_{\mathrm{HO}}
    \begin{bmatrix}
        \vec{b}_{\mathrm{p}, 1}
        \vec{b}_{\mathrm{p}, 2}
    \end{bmatrix} 
\end{equation}
as the uncertainty of the handover location distribution $\vec{p}_{\mathrm{HO}}$
in the orthogonal directions. The projected matrix $\vec{\Sigma}_{\mathrm{HO}, \phi}$
represents an ellipse with half axis $r_{1}$ and $r_{2}$ and angle $\theta$ from
the positive horizontal axis to $r_{1}$,
while the bounding of the orthogonal errors $e^{\bot}_{\mathrm{p, HO}, 1}$ and
$e^{\bot}_{\mathrm{p, HO}, 2}$ is performed by box constraints. Thus, it is
necessary to find the smallest axis-aligned rectangle that includes the ellipse
given by $r_{1}$, $r_{2}$, and $\theta$ in the coordinates with the basis vectors
$\vec{b}_{\mathrm{p, 1}}$ and $\vec{b}_{\mathrm{p, 2}}$.
This rectangle is given by
%\begin{subequations}
%    \begin{align}
%        x_{1, \mathrm{max}} &= \sqrt{r_{1}^{2} \cos^{2}(\theta) + r_{2}^{2} \sin^{2}(\theta)} \\
%        x_{2, \mathrm{max}} &= \sqrt{r_{1}^{2} \sin^{2}(\theta) + r_{2}^{2}
%        \cos^{2}(\theta)} \text{ ,}
%    \end{align}
%\end{subequations}
the maximum coordinates in the basis direction $\vec{b}_{\mathrm{p}, 1}$ and
$\vec{b}_{\mathrm{p}, 2}$, called $x_{1, \mathrm{max}}$ and $x_{2,
\mathrm{max}}$, respectively. Thus, the approximation of the handover location distribution
$\vec{p}_{\mathrm{HO}}$ in the path coordinates is defined by the box constraints
$(e^{\bot}_{\mathrm{p, HO}, 1} - x_{1, \mathrm{max}}, e^{\bot}_{\mathrm{p, HO}, 1} +
x_{1, \mathrm{max}})$ for the basis direction $\vec{b}_{\mathrm{p}, 1}$ and 
$(e^{\bot}_{\mathrm{p, HO}, 2} - x_{2, \mathrm{max}}, e^{\bot}_{\mathrm{p, HO}, 2} +
x_{2, \mathrm{max}})$ for the basis direction $\vec{b}_{\mathrm{p}, 2}$ at the path
parameter $\phi=\phi_{\mathrm{HO}}$. Note that the uncertainty in the path parameter
$\phi_{\mathrm{HO}}$ is not considered.

\subsubsection{Prediction Model Adaptions}
\label{sssec:prediction_adaptions}
The handover location becomes more confident as the robot converges towards the
human hand. Eventually, the prediction model~\cref{eq:prediction_model} will not
be needed anymore, and the measured human position $\vec{p}_{\mathrm{h}}$ is used
instead. The vector $\transpose{\vec{p}}_{\mathrm{h}, \phi} = [e_{\mathrm{p, h,
1}}^{\bot}, e_{\mathrm{p, h, 2}}^{\bot}, \phi_{\mathrm{h}}]$ represents the current human
position $\vec{p}_{\mathrm{h}}$ in the path system similar
to~\cref{eq:path_representation} and~\cref{eq:path_projection}.

The estimated handover location is given as the linear interpolation 
\begin{equation}
    \begin{aligned}
        \label{eq:goal_interpolation}
        \transpose{\tilde{\vec{p}}_{\mathrm{HO}, \phi}} &= 
        \begin{bmatrix}
            \tilde{e}^{\bot}_{\mathrm{p, HO}, 1}&
            \tilde{e}^{\bot}_{\mathrm{p, HO}, 2}&
            \tilde{\phi}_{\mathrm{HO}}
        \end{bmatrix} \\ &= 
            w_{\mathrm{pred}}(\alpha_{\mathrm{p}}, d_\mathrm{p}) 
            \transpose{\vec{\mu}_{\mathrm{HO}, \phi}} + (1 -
            w_{\mathrm{pred}}(\alpha_{\mathrm{p}}, d_\mathrm{p})) \transpose{\vec{p}_{\mathrm{h},
            \phi}}
    \end{aligned}
\end{equation}
between $\vec{p}_{\mathrm{h}, \phi}$ and $\vec{\mu}_{\mathrm{HO}, \phi}$ using the
weight $0 \leq w_{\mathrm{pred}} \leq 1$ with
\begin{equation}
    \label{eq:step_function}
    w_{\mathrm{pred}}(\alpha_{\mathrm{p}}, d_\mathrm{p}) = \frac{1}{2}  + \frac{1}{2} 
    \tanh\left(\alpha_{\mathrm{p}}d_{\mathrm{pred}}
    - d_\mathrm{p}\right)
\end{equation}
and 
$d_{\mathrm{pred}} = \phi_{\mathrm{h}} - \phi_{\mathrm{HO}}$
as the distance of the human hand from the handover location in path direction.
The parameters $\alpha_{\mathrm{p}}$ and $d_\mathrm{p}$ of the
function~\cref{eq:step_function} control the distances at which
the interpolation starts and stops.
%Equation~\cref{eq:goal_interpolation}
%is visualized for $\tilde{\phi}_{\mathrm{HO}}$ in~\cref{fig:goal_interpolation}.

Furthermore, as the human hand approaches the handover location, i.e.,
$d_{\mathrm{pred}} \to 0$, the uncertainty of the handover location becomes
smaller. Thus, the projected uncertainty values $x_{\mathrm{1, max}}$ and
$x_{\mathrm{2, max}}$ are set to zero for $w_{\mathrm{pred}} \approx 1$.
A linear interpolation, as in~\cref{eq:goal_interpolation}, is not
necessary in this case, as it is assumed that the uncertainty of the model
already decreases when $d_{\mathrm{pred}} \to 0$.

%\begin{figure}
%    \centering
%    \def\axisdefaultwidth{\linewidth}
%    \def\axisdefaultheight{0.4\linewidth}
%    \input{plots/goal_interpolation.tex}
%    \caption{Path parameter $\tilde{\phi}_{\mathrm{HO}}$ corresponding to the
%        handover goal position according to~\cref{eq:goal_interpolation}. The
%    horizontal axis is inverted.}
%    \label{fig:goal_interpolation}
%\end{figure}

\subsubsection{Orientation Prediction}
\label{sssec:orientation_prediction}

The orientation of the handover location is not part of the Gaussian process
model; instead, the current rotation matrix of the human hand
$\vec{R}_{\mathrm{h}}$ is used. Again, $\vec{R}_{\mathrm{h}}$ is given in the
global coordinate system and needs to be projected onto the path. Our
previous work~\citep{oelerichBoundMPCCartesianTrajectory2024} shows that projecting $\vec{R}_{\mathrm{h}}$ onto the path is
equivalent to computing the RPY angles in the coordinate system given by the
orientation basis vectors $\vec{b}_{\mathrm{o, 1}}$, $\vec{b}_{\mathrm{o, 2}}$,
and the normed angular velocity vector $\vec{b}_{\omega_\mathrm{r}}$ of the
last reference path segment. Using the projection matrix
\begin{equation}
    \vec{R}_{\mathrm{proj}} = \left[\begin{array}{c|c|c}
            \vec{b}_{\mathrm{o, 2}} &
            \vec{b}_{\omega_\mathrm{r}} &
            \vec{b}_{\mathrm{o, 1}} \text{~,}
    \end{array}\right]
\end{equation}
the rotation matrix $\vec{R}_{\mathrm{h}}$ is projected according to
\begin{equation}
    \vec{R}_{\mathrm{RPY}} = \transpose{\vec{R}_{\mathrm{proj}}}
    \vec{R}_{\mathrm{h}} \vec{R}_{\mathrm{proj}}
\end{equation}
into the coordinate system defined by
$\vec{R}_{\mathrm{proj}}$, from which the RPY angles are calculated.
Thus, the yaw, roll, and pitch angles of
$\vec{R}_{\mathrm{RPY}}$ are the orthogonal orientation errors
$e^{\bot}_{\mathrm{o, h, 1}}$ and $e^{\bot}_{\mathrm{o, h, 2}}$ and the
tangential orientation error $e_{\mathrm{o, h}}^{||}$ of $\vec{R}_{\mathrm{h}}$,
respectively.

Using this projection, the orientation goal w.r.t. the orientation path is given
by $e_{\mathrm{o, h, 1}}^{\bot}$ and $e_{\mathrm{o, h, 2}}^{\bot}$. These
orthogonal orientation errors are
adapted according to
\begin{equation}
    \tilde{e}_{\mathrm{o, h}, i}^{\bot} = 
    (1 - w_{\mathrm{pred}}(\alpha_{\mathrm{o}}, d_{\mathrm{o}})) e_{\mathrm{o, h},
        i}^{\bot}\text{~,~}i = 1, 2 \text{~,}
\end{equation}
with the parameters $\alpha_{\mathrm{o}}$ and $d_{\mathrm{o}}$, chosen similar
to $\alpha_{\mathrm{p}}$ and $d_{\mathrm{p}}$, controlling the
convergence of $\tilde{e}_{\mathrm{o, h}, i}^{\bot}$ towards
$e_{\mathrm{o, h}, i}^{\bot}$ as $d_{\mathrm{pred}} \to 0$. Thus, the desired
orientation errors $\tilde{e}_{\mathrm{o, h}, i}^{\bot}$ are approximately $0$ when the
robot is close to the approach point $\vec{p}_{\mathrm{r, a}}$ and arrives at the
projected orientation of the human hand
$e_{\mathrm{o, h}, i}^{\bot}$ at the handover location.
The handover location $\tilde{\vec{p}}_{\mathrm{HO}}$ is assumed to be close to
$\tilde{\vec{p}}_{\mathrm{HO, 0}}$ such that minimizing $e_{\mathrm{o, h}}^{||}$
leads to the desired approach orientation for $\tilde{\vec{p}}_{\mathrm{HO}}$.

%Similarly, the desired tangential orientation error $\tilde{e}_{\mathrm{o,
%r}}^{||}$ is set align the robot end-effector\textquotesingle s orientation with
%the approach direction between the approach point $\vec{p}_{\mathrm{r, a}}$ and
%the handover location $\tilde{\vec{p}}_{\mathrm{HO}}$ which is indicated in
%\cref{fig:handover_visualization}.

%The parallel orientation error $e_{\mathrm{o, r}}^{||}$ is given through the
%coupling of orientation and position path by the path parameter $\phi$.
%This ensures that the approach direction of the reference path set in
%\cref{ssec:reference_paths} is achieved. However, the reference path is set at
%the start of the handover and needs adaption during the handover. For this,
%$e_{\mathrm{o, r}}^{||}$ is used to correct the approach orientation. By
%choosing the normed reference angular velocity to be
%$\transpose{\vec{b}_{\omega_\mathrm{r}}} = [0, 0, 1]$, the parallel error
%$e_{\mathrm{o, r}}^{||}$ is used to make the gripper approach the handover
%location in line with its gripping direction. For this the desired parallel
%error
%\begin{equation}
%    \tilde{e}_{\mathrm{o, r}}^{||} = \frac{\pi}{2} +
%    \mathrm{arctan2}\left(\frac{p_{\mathrm{r, 2}} - p_{\mathrm{h,
%        2}}}{p_{\mathrm{r, 1}} - p_{\mathrm{h, 1}}}\right)
%\end{equation}
%with the $x$ and $y$ components $p_{\mathrm{r, 1}}$ and $p_{\mathrm{r, 2}}$ of
%the robot current position, respectively, and analogously of the human position.
%This aligns the robots gripping direction with the vector between the current
%human and robot position and, thus, ensures a successful handover.

\subsection{Motion Synchronization}
\label{ssec:desired_path_state}

The BoundMPC formulation includes a desired path state
$\transpose{\vec{\xi}}_{\mathrm{d}} = [\phi_{\mathrm{d}},
\dot{\phi}_{\mathrm{d}}, \ddot{\phi}_{\mathrm{d}}]$ in its objective
function~\cref{eq:mpc_objective},
which is used to control the path progress~\citep{oelerichBoundMPCCartesianTrajectory2024}. In this work,
$\vec{\xi}_{\mathrm{d}}$ enables the motion synchronization between the human
and the robot as required for a handover. Specifically, $\phi_{\mathrm{d}}$ is
combined with the error bounds in \cref{ssec:error_bounds} to reach
the handover location \goalconv, while  $\dot{\phi}_{\mathrm{d}}$ is used to
synchronize the human and the robot motion speed \goalsync. Setting 
\begin{equation}
    \label{eq:desired_path_param}
    \phi_{\mathrm{d}} = \tilde{\phi}_{\mathrm{HO}}
\end{equation}
ensures that the
robot moves as far along the path as is necessary to
reach the handover location. 

Synchronizing the path progress of the human and the robot brings along that the
handover location is reached approximately simultaneously by both actors. The
signed distances of the path parameter of the
current human hand position $\phi_{\mathrm{h}}$ and the current robot position
$\phi_{\mathrm{c}}$ to the handover point $\tilde{\phi}_{\mathrm{HO}}$
are
\begin{equation}
    \label{eq:dist_human_ho}
    d_{\mathrm{h, HO}}(t) = \phi_{\mathrm{h}}(t) - \tilde{\phi}_{\mathrm{HO}}
\end{equation}
and 
\begin{equation}
    \label{eq:dist_robot_ho}
    d_{\mathrm{r, HO}}(t) = \tilde{\phi}_{\mathrm{HO}} - \phi_{\mathrm{c}}(t) \text{ ,}
\end{equation}
respectively. The desired speed along the path is chosen as
\begin{equation}
    \label{eq:path_vel_d}
    \dot{\phi}_{\mathrm{d}} = \dot{\phi}_{\mathrm{d, max}} \tanh\left(s (d_{\mathrm{r, HO}}
    - d_{\mathrm{h, HO}} + b)\right) \text{ ,}
\end{equation}
with the maximum speed $\dot{\phi}_{\mathrm{d, max}}$ and the parameters $s$ and
$b$ influencing the shape of the smooth function~\cref{eq:path_vel_d}. Note
that~\cref{eq:path_vel_d} can be interpreted as a P controller with saturation.
The desired path velocity $\dot{\phi}_{\mathrm{d}}$ in~\cref{eq:path_vel_d} can
become negative if the human hand is further away from the handover location
than the robot by an offset $b$, incentivizing the robot to slow down or move
backwards
along the path. In classical path-following control, the path velocity must be
typically positive. This requirement is relaxed here to realize a more
natural human-robot interaction. 

\subsection{Error Bounds}
\label{ssec:error_bounds}

\begin{figure}
    \centering
    \def\axisdefaultwidth{0.9\linewidth}
    \def\axisdefaultheight{0.5\linewidth}
    \input{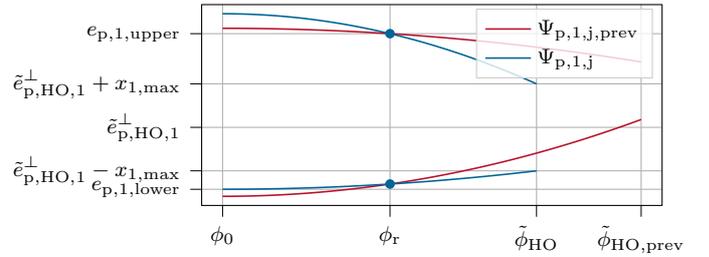}
    \caption{Error bounding functions for $e_{\mathrm{p, r}, 1}^{\bot}$ in two
        consecutive time steps where $\Psi_{\mathrm{p, l}, j, \mathrm{prev}}$
        and $\Psi_{\mathrm{p, l}, j}$ are the error functions in the first and
    second time step, respectively. The bounding functions range from the start
of the handover segment at $\phi_{0}$ to the handover location at
$\tilde{\phi}_{\mathrm{HO}}$. The previous estimation of the handover location
path parameter is $\tilde{\phi}_{\mathrm{HO, prev}}$}
    \label{fig:error_bounds}
\end{figure}

A successful handover requires the robot\textquotesingle s end-effector position
$\vec{p}_{\mathrm{r}}$ to converge to the position of the human hand
$\vec{p}_{\mathrm{h}}$ at the handover point $\tilde{\vec{p}}_{\mathrm{HO}}$
\goalconv.
Furthermore, it is required that the robot moves predictably such that
the human can anticipate the handover trajectory and does not collide with the
human hand (\goalpredictnb, \goalsafenb). 

The allowed deviations in orthogonal direction are determined by the error
bounding functions in the position $\Psi_{\mathrm{p}, i, j}(\phi)$ and the
orientation $\Psi_{\mathrm{o}, i, j}(\phi)$ for the basis directions $i = 1, 2$
and $j \in \{\mathrm{upper, lower}\}$. Thus, there are a total of eight error
bounding functions, i.e., one upper bound ($j = \mathrm{upper}$) and one lower
bound ($j = \mathrm{lower}$) for each orthogonal direction in the position and
orientation, which are updated in each time step.

Second-order functions serve as bounding functions. The replanning of the bounds
is based on the bounds from the previous time step $\Psi_{\mathrm{p}, i, j,
\mathrm{prev}}$. A visual explanation of the adaption for one orthogonal
direction is given in \textbf{\cref{fig:error_bounds}}. Specifically, the bounds
at the new time step coincide with the bounds of the previous time step at the
current path parameter $\phi_{\mathrm{r}}$ with the values $\Psi_{\mathrm{p}, i,
j}(\phi_{r}) = \Psi_{\mathrm{p}, i, j, \mathrm{prev}}(\phi_{r}) = e_{\mathrm{p}, i, j}$. The other
directions are adapted analogously.

The described adaption of the allowed orthogonal deviation from the reference
paths ensures that, as the robot traverses the path, the bounds converge to the
desired handover location \goalconv. As the bounds gradually narrow, the movement of the
robot\textquotesingle s end-effector becomes more restricted, ensuring that no
sudden movements occur and the motion stays predictable \goalpredict. Furthermore, the
desired path parameter $\phi_{\mathrm{d}} = \tilde{\phi}_{\mathrm{HO}}$
from~\cref{eq:desired_path_param}, in combination with the error bounds, ensures
that the robot\textquotesingle s end-effector does not collide with the human
hand as the end-effector is incentivized to stay in front of the human hand
\goalsafe.

\subsection{Terminal Cost}
\label{ssec:terminal_cost}

With the specified error bounds from~\cref{ssec:error_bounds}, the robot is
incentivized to move to the goal. In order to further improve the goal-seeking behavior,
a terminal cost term is added to the MPC formulation~\cref{eq:mpc_problem} \goalconv. A reasonable choice
is the tracking cost 
\begin{equation}
    \label{eq:terminal_cost}
    J_{\mathrm{T}} = w_{\mathrm{T, p}} \norm{
        \begin{bmatrix}
            e_{\mathrm{p, r, 1}}^{\bot} - \tilde{e}^{\bot}_{\mathrm{p, HO, 1}} \\
            e_{\mathrm{p, r, 2}}^{\bot} -
        \tilde{e}^{\bot}_{\mathrm{p, HO, 2}}
        \end{bmatrix}
    }^{2}_{2} + \, w_{\mathrm{T, o}} \norm{
        \begin{bmatrix}
            e^{\bot}_{\mathrm{o, r, 1}} - \tilde{e}^{\bot}_{\mathrm{o, h, 1}} \\
            e^{\bot}_{\mathrm{o, r, 2}} - \tilde{e}^{\bot}_{\mathrm{o, h, 2}}
            %e^{||}_{\mathrm{o, r}} - \tilde{e}^{||}_{\mathrm{o, r}}
        \end{bmatrix}
    }^{2}_{2}
\end{equation}
with the positive weights $w_{\mathrm{T, p}}$ and $w_{\mathrm{T, o}}$. Extending the
cost~\cref{eq:mpc_objective} by~\cref{eq:terminal_cost} makes the
robot\textquotesingle s end-effector converge to the adapted handover
location $\tilde{\vec{p}}_{\mathrm{HO}}$.

\section{Handover Experiment}

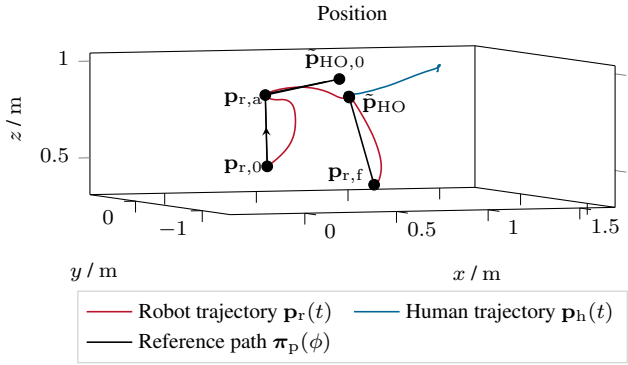
\begin{figure}
    \centering
    \def\axisdefaultwidth{\linewidth}
    \def\axisdefaultheight{0.45\linewidth}
    \begin{tikzpicture}
\definecolor{darkgray176}{RGB}{176,176,176}
\definecolor{darkorange25512714}{RGB}{0,102,153}
\definecolor{acinyellow}{RGB}{252, 204, 71}
\definecolor{forestgreen4416044}{RGB}{0,190,65}
\definecolor{steelblue31119180}{RGB}{186,18,43}
\definecolor{lightgray204}{RGB}{204,204,204}
\begin{axis}[
    title=Position,
    view={-20}{3},
    xlabel={$x$ / \si{\meter}},
    ylabel={$y$ / \si{\meter}},
    zlabel={$z$ / \si{\meter}},
    axis equal,
    tick align=outside,
    tick pos=left,
    x grid style={darkgray176},
    y grid style={darkgray176},
    xtick style={color=black},
    ytick style={color=black},
    legend cell align={left},
    legend columns=2,
    legend style={
      fill opacity=0.8,
      draw opacity=1,
      anchor=north,
      at={(0.5, -0.45)},
      text opacity=1,
      draw=lightgray204
    },
    legend entries ={Robot trajectory $\vec{p}_{\mathrm{r}}(t)$, Human
    trajectory $\vec{p}_{\mathrm{h}}(t)$, Reference path
$\vec{\pi}_{\mathrm{p}}(\phi)$}
    ]
\pgfplotstableread{plots/traj_robot_simple.txt}\trajrobot;
\pgfplotstableread{plots/traj_human_simple.txt}\trajhuman;
\pgfplotstableread{plots/traj_pred.txt}\trajpred;
\pgfplotstableread{plots/via_points_simple.txt}\vias;
    \addplot3 [steelblue31119180, semithick]
            table [
               x expr=\thisrowno{0}, 
               y expr=\thisrowno{1}, 
               z expr=\thisrowno{2} 
             ] {\trajrobot};
    \addplot3 [darkorange25512714, semithick]
            table [
               x expr=\thisrowno{0}, 
               y expr=\thisrowno{1}, 
               z expr=\thisrowno{2} 
             ] {\trajhuman};
    \addplot3 [black, semithick, skip coords between index={50}{1000}]
            table [
               x expr=\thisrowno{6}, 
               y expr=\thisrowno{7}, 
               z expr=\thisrowno{8} 
             ] {\trajrobot};
    \addplot3 [black, semithick]
            table [
               x expr=\thisrowno{0}, 
               y expr=\thisrowno{1}, 
               z expr=\thisrowno{2} 
             ] {\trajpred};
     \addplot3 [black, mark=*, only marks, skip coords between index={0}{1}]
        table [
           x expr=\thisrowno{0}, 
           y expr=\thisrowno{1}, 
           z expr=\thisrowno{2} 
         ] {\trajpred};
    \addplot3 [black, semithick, skip coords between index={0}{70}]
            table [
               x expr=\thisrowno{6}, 
               y expr=\thisrowno{7}, 
               z expr=\thisrowno{8} 
             ] {\trajrobot};
    \addplot3 [black, mark=*, only marks, skip coords between index={72}{1000}, skip coords between index={0}{70}]
            table [
               x expr=\thisrowno{6}, 
               y expr=\thisrowno{7}, 
               z expr=\thisrowno{8} 
             ] {\trajrobot};
     \addplot3 [black, mark=*, only marks, skip coords between index={2}{3}]
        table [
           x expr=\thisrowno{0}, 
           y expr=\thisrowno{1}, 
           z expr=\thisrowno{2} 
         ] {\vias};
    \addplot3 [black, postaction={decorate}, decoration={markings, mark=at
        position 0.55 with {\arrow{stealth}}}, skip coords between index={2}{100}]
            table [
               x expr=\thisrowno{0}, 
               y expr=\thisrowno{1}, 
               z expr=\thisrowno{2} 
             ] {\vias};

\node[] at (axis cs: 0.0, -0.5, 0.5) {$\vec{p}_{\mathrm{r, 0}}$};
\node[] at (axis cs: 0.0, -0.5, 0.85) {$\vec{p}_{\mathrm{r, a}}$};
\node[] at (axis cs: 0.5, -0.5, 1.05) {$\tilde{\vec{p}}_{\mathrm{HO, 0}}$};
\node[] at (axis cs: 0.77, -0.5, 0.82) {$\tilde{\vec{p}}_{\mathrm{HO}}$};
\node[] at (axis cs: 0.56, -0.5, 0.45) {$\vec{p}_{\mathrm{r, f}}$};

\end{axis}
\end{tikzpicture}
    \caption{Position trajectories of the robot\textquotesingle s end-effector
    $\vec{p}_{\mathrm{r}}(t)$ and human hand $\vec{p}_{\mathrm{h}}(t)$.}
    \label{fig:human_robot_simple_3d}
\end{figure}

The performance of the developed MPC formulation is demonstrated by a
human-to-robot handover experiment where a human hands a cup to the robot. The trajectories
of the human hand and the robot\textquotesingle s end effector are shown in
\textbf{\cref{fig:human_robot_simple_3d}}. The robot moves from an
initial pose $\vec{p}_{\mathrm{r, 0}}$ to an approach point $\vec{p}_{\mathrm{r,
a}}$ from which it continues to perform the handover at
$\tilde{\vec{p}}_{\mathrm{HO}}$. The reference path is initially planned
towards the first prediction of the handover location
$\tilde{\vec{p}}_{\mathrm{HO, 0}}$. However, the robot deviates from
this path to reach the actual handover location $\tilde{\vec{p}}_{\mathrm{HO}}$,
computed during the handover. Afterwards, a new reference path segment
is planned from $\tilde{\vec{p}}_{\mathrm{HO}}$ to the final pose
$\vec{p}_{\mathrm{r, f}}$ at which the robot puts the object down. 
%No error bound adaption is needed up to
%the approach point $\vec{p}_{\mathrm{r, a}}$ since it is fixed. Also, the
%terminal weights $w_{\mathrm{T, p}}$ and $w_{\mathrm{T, o}}$ are $0$ until the
%robot reaches the approach point $\vec{p}_{\mathrm{r, a}}$.
Using OptiTrack, The cup is tracked instead of the human hand since it
represents the object of interest. A video of the experiments is available at
\url{www.acin.tuwien.ac.at/42d1/}.

\begin{figure}
    \centering
    \def\axisdefaultwidth{\linewidth}
    \def\axisdefaultheight{0.38\linewidth}
    % This file was created with tikzplotlib v0.10.1.
\begin{tikzpicture}

\definecolor{darkgray176}{RGB}{176,176,176}
\definecolor{firebrick1861843}{RGB}{186,18,43}
\definecolor{lightgray204}{RGB}{204,204,204}
\definecolor{sandybrown25220370}{RGB}{252,203,70}
\definecolor{teal0101153}{RGB}{0,101,153}

\begin{axis}[
legend cell align={left},
legend style={
  fill opacity=0.8,
  draw opacity=1,
  text opacity=1,
  at={(0.97,0.03)},
  anchor=south east,
  draw=lightgray204
},
tick align=outside,
tick pos=left,
x grid style={darkgray176},
xmin=1.10000001600019, xmax=5.40000008000015,
xtick style={color=black},
xtick={2,3,4,5},
xticklabels={\empty,\empty,\empty,\empty},
y grid style={darkgray176},
ylabel={\(\displaystyle \phi\) / \si{\meter}},
ymin=0.344632926220438, ymax=2.02311461833877,
ytick style={color=black},
ytick={0.4,0.8,1.2,1.6,2},
yticklabels={
  \(\displaystyle {0.4}\),
  \(\displaystyle {0.8}\),
  \(\displaystyle {1.2}\),
  \(\displaystyle {1.6}\),
  \(\displaystyle {2.0}\)
}
]
\addplot [semithick, firebrick1861843]
table {%
1.10000001600019 0.420927548589453
1.20000001700009 0.466840885652748
1.30000001899998 0.511593254285533
1.40000002000033 0.55549553998525
1.50000002200022 0.598955372110497
1.60000002300012 0.641921581508374
1.70000002500001 0.683963486716232
1.80000002600036 0.724320273010643
1.90000002800025 0.762096072696227
2.00000002900015 0.796535431044647
2.10000003100004 0.82716082545183
2.20000003199993 0.854055986602648
2.30000003400028 0.87757662302353
2.40000003500018 0.898643461241185
2.50000003700006 0.917217004727121
2.60000003799996 0.933614087213972
2.7000000400003 0.949893046580637
2.8000000410002 0.967635951996897
2.90000004300009 0.988608827767567
3.00000004399999 1.01463402653193
3.10000004600033 1.04605256691528
3.20000004700023 1.08091722190628
3.30000004900012 1.11674966634526
3.40000005000002 1.15218862208419
3.50000005200036 1.18633145277741
3.60000005300026 1.21797367118742
3.70000005500015 1.2457704383675
3.80000005600004 1.2691034498334
3.90000005799993 1.28810411516077
4.00000005900029 1.30295711395361
4.10000006100017 1.31420782412686
4.20000006200007 1.32235219556582
4.30000006399996 1.32780582258131
4.40000006500031 1.33188698682717
4.5000000670002 1.3355688964639
4.6000000680001 1.338938110203
4.70000006999999 1.3419410122177
4.80000007100034 1.34436996005273
4.90000007300023 1.34595279640456
5.00000007400013 1.34668490890424
5.10000007600001 1.34683128977999
5.20000007700037 1.34658772555612
5.30000007800027 1.34610175622539
5.40000008000015 1.34549948225141
};
\addlegendentry{$\phi_\mathrm{c}$}
\addplot [semithick, teal0101153]
table {%
1.10000001600019 1.93922996748819
1.20000001700009 1.94681999596975
1.30000001899998 1.94604154346697
1.40000002000033 1.94300116353242
1.50000002200022 1.93782616056706
1.60000002300012 1.93449967735615
1.70000002500001 1.93276607637819
1.80000002600036 1.93328864118968
1.90000002800025 1.9371794552569
2.00000002900015 1.94051909475082
2.10000003100004 1.93611099564587
2.20000003199993 1.92286588440841
2.30000003400028 1.8894500212927
2.40000003500018 1.82572819559982
2.50000003700006 1.74685735754964
2.60000003799996 1.67446418577389
2.7000000400003 1.61409432824706
2.8000000410002 1.55949250535479
2.90000004300009 1.51020492321757
3.00000004399999 1.46802339275711
3.10000004600033 1.4345918484907
3.20000004700023 1.40734686302123
3.30000004900012 1.38781435369912
3.40000005000002 1.37355262272613
3.50000005200036 1.36380408557496
3.60000005300026 1.35925156697626
3.70000005500015 1.35742910242783
3.80000005600004 1.35802458316397
3.90000005799993 1.35917951736356
4.00000005900029 1.35829145161431
4.10000006100017 1.35358771682487
4.20000006200007 1.34969155632458
4.30000006399996 1.34912016992718
4.40000006500031 1.35234906832375
4.5000000670002 1.3515227988103
4.6000000680001 1.34980019935854
4.70000006999999 1.34803918266601
4.80000007100034 1.34503616139978
4.90000007300023 1.34529860561297
5.00000007400013 1.34490726308762
5.10000007600001 1.34391435141547
5.20000007700037 1.3436964929465
5.30000007800027 1.34291491629233
5.40000008000015 1.34316193909459
};
\addlegendentry{$\phi_\mathrm{h}$}
\addplot [semithick, sandybrown25220370]
table {%
1.10000001600019 1.42964269070586
1.20000001700009 1.43059721346225
1.30000001899998 1.41805769700889
1.40000002000033 1.41915295843351
1.50000002200022 1.41695075122613
1.60000002300012 1.41527853845726
1.70000002500001 1.41696558688081
1.80000002600036 1.42201899599261
1.90000002800025 1.43162471010398
2.00000002900015 1.43597171984093
2.10000003100004 1.44355836210484
2.20000003199993 1.43565738212174
2.30000003400028 1.43779603734743
2.40000003500018 1.39154530326066
2.50000003700006 1.38420333129652
2.60000003799996 1.36394292652485
2.7000000400003 1.34299675181614
2.8000000410002 1.3619204665526
2.90000004300009 1.41595586504061
3.00000004399999 1.41880401572488
3.10000004600033 1.42595190255595
3.20000004700023 1.40534653127745
3.30000004900012 1.38755340643695
3.40000005000002 1.37300144488685
3.50000005200036 1.36357625679141
3.60000005300026 1.36044702847441
3.70000005500015 1.35911538192625
3.80000005600004 1.36018515715848
3.90000005799993 1.36115408809832
4.00000005900029 1.35866550007497
4.10000006100017 1.35354552747419
4.20000006200007 1.34997530369137
4.30000006399996 1.35055382092407
4.40000006500031 1.35318553745938
4.5000000670002 1.35206944992327
4.6000000680001 1.35074360718396
4.70000006999999 1.34940926015838
4.80000007100034 1.34475651486171
4.90000007300023 1.34544687008668
5.00000007400013 1.34490898404066
5.10000007600001 1.34381818427095
5.20000007700037 1.34363858553709
5.30000007800027 1.34415818796699
5.40000008000015 1.34326668901031
};
\addlegendentry{$\tilde{\phi}_\mathrm{HO}$}
\end{axis}

\end{tikzpicture}
    \def\axisdefaultheight{0.3\linewidth}
    % This file was created with tikzplotlib v0.10.1.
\begin{tikzpicture}

\definecolor{darkgray176}{RGB}{176,176,176}
\definecolor{firebrick1861843}{RGB}{186,18,43}

\begin{axis}[
tick align=outside,
tick pos=left,
x grid style={darkgray176},
xlabel={\(\displaystyle t\) / \si{\second}},
xmin=1.10000001600019, xmax=5.40000008000015,
xtick style={color=black},
xtick={2,3,4,5},
xticklabels={
  \(\displaystyle {2}\),
  \(\displaystyle {3}\),
  \(\displaystyle {4}\),
  \(\displaystyle {5}\)
},
y grid style={darkgray176},
ylabel={\(\displaystyle \dot{q}_2\) / \si{\degree\per\second}},
ymin=-56, ymax=56,
ytick style={color=black},
ytick={-55,0,55},
yticklabels={
  \(\displaystyle \underline{\dot{q}_2}\),
  0,
  \(\displaystyle \overline{\dot{q}_2}\)
}
]
\addplot [semithick, firebrick1861843]
table {%
1.10000001600019 -40.4179452454683
1.20000001700009 -38.3596113154469
1.30000001899998 -36.3100054715024
1.40000002000033 -34.5671770468596
1.50000002200022 -33.1160861244783
1.60000002300012 -31.8509722011564
1.70000002500001 -30.7082892554544
1.80000002600036 -29.5713188147941
1.90000002800025 -28.3100496823776
2.00000002900015 -26.8951880117836
2.10000003100004 -25.7131529651024
2.20000003199993 -24.704204892051
2.30000003400028 -23.232379677959
2.40000003500018 -21.1941443875864
2.50000003700006 -18.1438298569092
2.60000003799996 -14.557175679719
2.7000000400003 -10.6987558430317
2.8000000410002 -5.09336952160115
2.90000004300009 3.44788361940842
3.00000004399999 17.2855455390199
3.10000004600033 31.3817555874111
3.20000004700023 42.7509677856361
3.30000004900012 50.5714390019909
3.40000005000002 54.2464142679203
3.50000005200036 55.0000005727768
3.60000005300026 54.8704421411961
3.70000005500015 54.0066705924976
3.80000005600004 52.1007537978711
3.90000005799993 49.3715224346337
4.00000005900029 44.9647345720133
4.10000006100017 37.7004611362349
4.20000006200007 28.6501184614034
4.30000006399996 19.8855975107882
4.40000006500031 12.8145606457675
4.5000000670002 7.61823623117584
4.6000000680001 4.77988031228629
4.70000006999999 3.82795315057349
4.80000007100034 3.34085050829956
4.90000007300023 2.65160889893295
5.00000007400013 1.84343604942689
5.10000007600001 1.05152028273783
5.20000007700037 0.383799966837205
5.30000007800027 -0.0447969318221853
5.40000008000015 -0.210776520326267
};
\end{axis}

\end{tikzpicture}
    \caption{Path parameter of the robot, the human, and the predicted handover
        location. Additionally, the joint velocity $\dot{q}_2$ of
        axis 2 is shown with the velocity constraints $\underline{\dot{q}_2}$
    and $\overline{\dot{q}_2}$.}
    \label{fig:path_param_simple}
\end{figure}
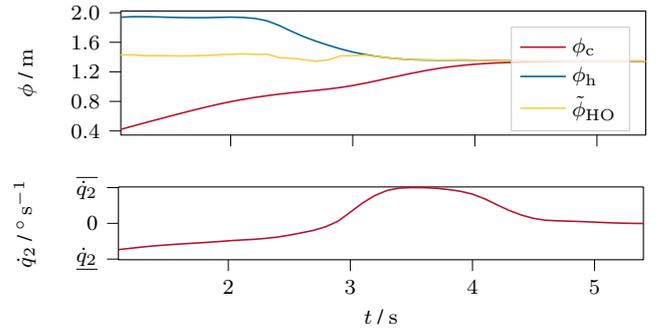

\textbf{\Cref{fig:path_param_simple}} shows the path parameter of the robot
$\phi_{\mathrm{r}}$, the human $\phi_{\mathrm{h}}$, and the predicted handover
location $\tilde{\phi}_{\mathrm{HO}}$. 
The robot tries to synchronize this motion, which results in an approximately
simultaneous arrival at the handover location $\tilde{\vec{p}}_{\mathrm{HO}}$.
However, the robot arrives slightly later due to the bound in the joint velocity
$\dot{q}_2$
of axis 2, which limits the robot\textquotesingle s end-effector velocity.
This also shows that the kinematic limits are exploited but not violated.
Furthermore, the robot slows down during $\SI{2}{\second} < t < \SI{2.5}{\second}$
since the human has not started moving yet. For $t > \SI{3}{\second}$, the
handover path location approaches $\tilde{\phi}_{\mathrm{HO}} \approx
\phi_{\mathrm{h}}$ due to the adaption in~\cref{eq:goal_interpolation}.

\begin{figure}
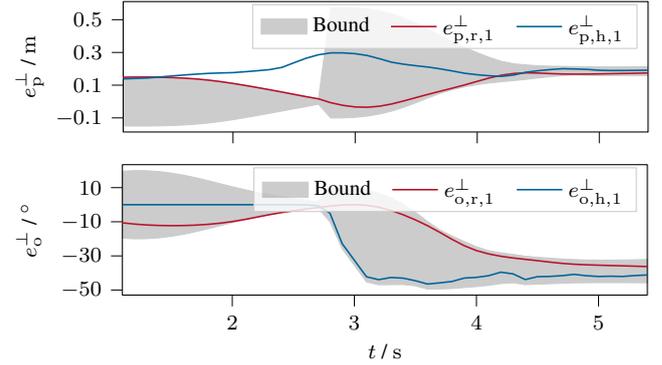

    \centering
    \def\axisdefaultwidth{\linewidth}
    \def\axisdefaultheight{0.39\linewidth}
    \input{plots/e_p_orth1_time.tex}
    \input{plots/e_r_orth0_time.tex}
    \caption{Orthogonal position errors $e_{\mathrm{p, r}, 1}^{\bot}$ and
        $e_{\mathrm{p, h}, 1}^{\bot}$ and orientation errors $e_{\mathrm{o, r},
        1}^{\bot}$ and $e_{\mathrm{o, h}, 1}^{\bot}$ during the handover. Only
    the part of the trajectory relevant to the handover is shown. The robot
errors $e_{\mathrm{p, r}, 1}^{\bot}$ and $e_{\mathrm{o, r}, 1}^{\bot}$ remain
within the bounds while the human errors $e_{\mathrm{p, h}, 1}^{\bot}$ and
$e_{\mathrm{o, h}, 1}^{\bot}$ are unbounded.}
    \label{fig:e_p_orth1}
\end{figure}

Further insights can be gained from the  orthogonal position errors
$e_{\mathrm{p, r, 1}}^{\bot}$ and $e_{\mathrm{p, h,
1}}^{\bot}$ and orientation errors $e_{\mathrm{o, r, 1}}^{\bot}$ and
$e_{\mathrm{o, h, 1}}^{\bot}$ shown in \textbf{\cref{fig:e_p_orth1}}. The robot
successfully traverses the via-point and enters the segment of the handover
where the bounds are adapted to converge to the position of the human hand. The
bounds are initially high due to the uncertainty of $\vec{p}_{\mathrm{HO}}$ and
then narrow as the robot approaches the human hand. Note that a low
orientation error is present at the final handover pose
$\tilde{\vec{p}}_{\mathrm{HO}}$ due to the weights $w_{\mathrm{T, p}} >
w_{\mathrm{T, o}}$. This low error still allows reliable human-to-robot
handovers.

%The bounds are determined by the uncertainty matrix $\vec{\Sigma}_{\mathrm{HO}}$
%of the prediction model. A simple measure of the uncertainty is the largest
%eigenvalue $\lambda_{\mathrm{max}}$ of $\vec{\Sigma}_{\mathrm{HO}}$ which is
%shown in~\cref{fig:prediction_max_eigenvalue}. As the human hand starts moving,
%the prediction model becomes less certain at first but improves again as the
%human approaches the handover location.

%\begin{figure}
%    \centering
%    \def\axisdefaultwidth{\linewidth}
%    \def\axisdefaultheight{0.4\linewidth}
%    \input{plots/e_r_orth0_time.tex}
%    %\input{plots/e_r_orth1_time.tex}
%    \caption{Temporal evolution of $e_{\mathrm{o, r}, 1}^{\bot}$ and
%        $e_{\mathrm{o, h}, 1}^{\bot}$ during the handover. Only
%    the part of the trajectory is shown that is relevant to the handover.}
%    \label{fig:e_o_orth1}
%\end{figure}

%\begin{figure}
%    \centering
%    \def\axisdefaultwidth{\linewidth}
%    \def\axisdefaultheight{0.35\linewidth}
%    \input{plots/prediction_max_eigenvalue.tex}
%    \caption{Largest eigenvalue $\lambda_{\mathrm{max}}$ of the prediction
%    uncertainty matrix $\vec{\Sigma}_{\mathrm{HO}}$ during the handover.}
%    \label{fig:prediction_max_eigenvalue}
%\end{figure}

\section{Conclusion}

This work proposes a novel trajectory planner for human-robot handovers. The
planner is based on a path-following model predictive control (MPC) formulation
where specific requirements for successful handovers are naturally encoded. A
handover location prediction is developed based on Gaussian process regression,
which predicts the final pose of the handover. This pose is then approached by
projecting it into the path system and using a terminal cost in combination with
Cartesian error-bound adaptions. These adaptions are made in each step to ensure
that the robot reacts appropriately to human motion. Finally, the feasibility of
the proposed approach is demonstrated on a 7-DoF robotic manipulator, which
receives an object from a human during a handover. Future work will deal with
improvements to human motion prediction and applying the developed
framework to other scenarios with dynamically changing environments and goals.

\printbibliography %Prints bibliography

@inproceedings{ardakaniRealtimeTrajectoryGeneration2015,
  title = {Real-Time Trajectory Generation Using Model Predictive Control},
  booktitle = {Proceedings of the {{IEEE International Conference}} on {{Automation Science}} and {{Engineering}} ({{CASE}})},
  author = {Ardakani, M. Mahdi Ghazaei and Olofsson, Bjorn and Robertsson, Anders and Johansson, Rolf},
  year = {2015},
  month = aug,
  pages = {942--948},
  publisher = {{IEEE}},
  address = {{Gothenburg, Sweden}},
  doi = {10.1109/CoASE.2015.7294220},
  urldate = {2023-03-31},
  isbn = {978-1-4673-8183-3},
  langid = {english}
}

@inproceedings{astudilloVaryingRadiusTunnelFollowingNMPC2022,
  title = {Varying-{{Radius Tunnel-Following NMPC}} for {{Robot Manipulators Using Sequential Convex Quadratic Programming}}},
  booktitle = {Proceedings of the {{Modeling}}, {{Estimation}} and {{Control Conference}}},
  author = {Astudillo, Alejandro and Pipeleers, Goele and Gillis, Joris and Decr{\'e}, Wilm and Swevers, Jan},
  year = {2022},
  volume = {55},
  pages = {345--352},
  address = {{Jersey City, USA}},
  doi = {10.1016/j.ifacol.2022.11.208},
  langid = {english}
}

@article{bethgeModelPredictiveControl2023,
  title = {Model {{Predictive Control}} with {{Gaussian-Process-Supported Dynamical Constraints}} for {{Autonomous Vehicles}}},
  author = {Bethge, Johanna and Pfefferkorn, Maik and Rose, Alexander and Peters, Jan and Findeisen, Rolf},
  year = {2023},
  month = jan,
  journal = {IFAC-PapersOnLine},
  series = {22nd {{IFAC World Congress}}},
  volume = {56},
  number = {2},
  pages = {507--512},
  doi = {10.1016/j.ifacol.2023.10.1618},
  urldate = {2023-12-20}
}

@inproceedings{carvalhoMotionPlanningDiffusion2023,
  title = {Motion {{Planning Diffusion}}: {{Learning}} and {{Planning}} of {{Robot Motions}} with {{Diffusion Models}}},
  shorttitle = {Motion {{Planning Diffusion}}},
  booktitle = {Proceedings of the {{IEEE}}/{{RSJ International Conference}} on {{Intelligent Robots}} and {{Systems}} ({{IROS}})},
  author = {Carvalho, Joao and Le, An T. and Baierl, Mark and Koert, Dorothea and Peters, Jan},
  year = {2023},
  month = oct,
  pages = {1916--1923},
  doi = {10.1109/IROS55552.2023.10342382},
  urldate = {2023-12-20}
}

@inproceedings{corsiniNonlinearModelPredictive2022,
  title = {Nonlinear {{Model Predictive Control}} for {{Human-Robot Handover}} with {{Application}} to the {{Aerial Case}}},
  booktitle = {Proceedings of the {{IEEE}}/{{RSJ International Conference}} on {{Intelligent Robots}} and {{Systems}} ({{IROS}})},
  author = {Corsini, Gianluca and Jacquet, Martin and Das, Hemjyoti and Afifi, Amr and Sidobre, Daniel and Franchi, Antonio},
  year = {2022},
  month = oct,
  pages = {7597--7604},
  issn = {2153-0866},
  doi = {10.1109/IROS47612.2022.9981045},
  urldate = {2023-10-24}
}

@article{elbanhawiSamplingBasedRobotMotion2014,
  title = {Sampling-{{Based Robot Motion Planning}}: {{A Review}}},
  shorttitle = {Sampling-{{Based Robot Motion Planning}}},
  author = {Elbanhawi, Mohamed and Simic, Milan},
  year = {2014},
  journal = {IEEE Access},
  volume = {2},
  pages = {56--77},
  doi = {10.1109/ACCESS.2014.2302442}
}

@article{hartl-nesicSurfacebasedPathFollowing2021,
  title = {Surface-Based Path Following Control: {{Application}} of Curved Tapes on 3-{{D}} Objects},
  author = {{Hartl-Nesic}, Christian and Gl{\"u}ck, Tobias and Kugi, Andreas},
  year = {2021},
  journal = {IEEE Transactions on Robotics},
  volume = {37},
  number = {2},
  pages = {615--626},
  doi = {10.1109/TRO.2020.3033721}
}

@article{hewingCautiousModelPredictive2020,
  title = {Cautious {{Model Predictive Control Using Gaussian Process Regression}}},
  author = {Hewing, Lukas and Kabzan, Juraj and Zeilinger, Melanie N.},
  year = {2020},
  journal = {IEEE Transactions on Control Systems Technology},
  volume = {28},
  number = {6},
  eprint = {1705.10702},
  pages = {2736--2743},
  publisher = {{IEEE}},
  issn = {15580865},
  doi = {10.1109/TCST.2019.2949757},
  archiveprefix = {arxiv}
}

@inproceedings{jankowskiVPSTOViapointbasedStochastic2023,
  title = {{{VP-STO}}: {{Via-point-based Stochastic Trajectory Optimization}} for {{Reactive Robot Behavior}}},
  shorttitle = {{{VP-STO}}},
  booktitle = {Proceedings of the {{IEEE International Conference}} on {{Robotics}} and {{Automation}} ({{ICRA}})},
  author = {Jankowski, Julius and Bruderm{\"u}ller, Lara and Hawes, Nick and Calinon, Sylvain},
  year = {2023},
  month = mar,
  eprint = {2210.04067},
  primaryclass = {cs},
  pages = {10125--10131},
  doi = {10.1109/ICRA48891.2023.10160214},
  urldate = {2023-08-01},
  archiveprefix = {arxiv}
}

@inproceedings{kshirsagarTimingSpecifiedControllersFeedback2022,
  title = {Timing-{{Specified Controllers}} with {{Feedback}} for {{Human-Robot Handovers}}},
  booktitle = {Proceedings of the {{IEEE International Conference}} on {{Robot}} and {{Human Interactive Communication}} ({{RO-MAN}})},
  author = {Kshirsagar, Alap and Ravi, Rahul Kumar and {Kress-Gazit}, Hadas and Hoffman, Guy},
  year = {2022},
  month = aug,
  pages = {1313--1320},
  issn = {1944-9437},
  doi = {10.1109/RO-MAN53752.2022.9900856},
  urldate = {2023-10-24}
}

@inproceedings{liProvablySafeEfficient2021,
  title = {Provably {{Safe}} and {{Efficient Motion Planning}} with {{Uncertain Human Dynamics}}},
  booktitle = {Proceedings of {{Robotics}}: {{Science}} and {{Systems}}},
  author = {Li, Shen and Figueroa, Nadia and Shah, Ankit and Shah, Julie},
  year = {2021},
  address = {{Virtual}},
  doi = {10.15607/rss.2021.xvii.050}
}

@article{lyu3DHumanMotion2022,
  title = {{{3D}} Human Motion Prediction: {{A}} Survey},
  shorttitle = {{{3D}} Human Motion Prediction},
  author = {Lyu, Kedi and Chen, Haipeng and Liu, Zhenguang and Zhang, Beiqi and Wang, Ruili},
  year = {2022},
  month = jun,
  journal = {Neurocomputing},
  volume = {489},
  pages = {345--365},
  issn = {09252312},
  doi = {10.1016/j.neucom.2022.02.045},
  urldate = {2022-11-25},
  langid = {english}
}

@misc{oelerichBoundMPCCartesianTrajectory2024,
  title = {{{BoundMPC}}: {{Cartesian Trajectory Planning}} with {{Error Bounds}} Based on {{Model Predictive Control}} in the {{Joint Space}}},
  shorttitle = {{{BoundMPC}}},
  author = {Oelerich, Thies and Beck, Florian and {Hartl-Nesic}, Christian and Kugi, Andreas},
  year = {2024},
  month = jan,
  number = {arXiv:2401.05057},
  eprint = {2401.05057},
  primaryclass = {cs},
  publisher = {{arXiv}},
  urldate = {2024-01-11},
  archiveprefix = {arxiv}
}

@article{osaMotionPlanningLearning2022,
  title = {Motion Planning by Learning the Solution Manifold in Trajectory Optimization},
  author = {Osa, Takayuki},
  year = {2022},
  month = mar,
  journal = {The International Journal of Robotics Research},
  volume = {41},
  number = {3},
  pages = {281--311},
  publisher = {{SAGE Publications Ltd STM}},
  issn = {0278-3649},
  doi = {10.1177/02783649211044405},
  urldate = {2023-08-17},
  langid = {english}
}

@article{perssonSamplingbasedAlgorithmRobot2014,
  title = {Sampling-Based {{A}}* Algorithm for Robot Path-Planning},
  author = {Persson, Sven Mikael and Sharf, Inna},
  year = {2014},
  journal = {The International Journal of Robotics Research},
  volume = {33},
  number = {13},
  pages = {1683--1708},
  doi = {10.1177/0278364914547786},
  langid = {english}
}

@book{rasmussenGaussianProcessesMachine2006,
  title = {Gaussian {{Processes}} for {{Machine Learning}}},
  author = {Rasmussen, Carl Edward and Williams, Christopher K. I.},
  year = {2006},
  series = {Adaptive Computation and Machine Learning},
  publisher = {{MIT Press}},
  address = {{Cambridge, Massachusetts}},
  isbn = {978-0-262-18253-9},
  langid = {english},
  lccn = {QA274.4 .R37 2006}
}

@article{schoelsCIAOMPCbasedSafe2020,
  title = {{{CIAO}}*: {{MPC-based}} Safe Motion Planning in Predictable Dynamic Environments},
  author = {Schoels, Tobias and Rutquist, Per and Palmieri, Luigi and Zanelli, Andrea and Arras, Kai O. and Diehl, Moritz},
  year = {2020},
  journal = {IFAC-PapersOnLine},
  volume = {53},
  number = {2},
  pages = {6555--6562},
  issn = {24058963},
  doi = {10.1016/j.ifacol.2020.12.072}
}

@article{schulmanMotionPlanningSequential2014,
  title = {Motion Planning with Sequential Convex Optimization and Convex Collision Checking},
  author = {Schulman, John and Duan, Yan and Ho, Jonathan and Lee, Alex and Awwal, Ibrahim and Bradlow, Henry and Pan, Jia and Patil, Sachin and Goldberg, Ken and Abbeel, Pieter},
  year = {2014},
  journal = {The International Journal of Robotics Research},
  volume = {33},
  number = {9},
  pages = {1251--1270},
  publisher = {{SAGE Publications Ltd STM}},
  doi = {10.1177/0278364914528132},
  langid = {english}
}

@misc{solaMicroLieTheory2021,
  title = {A Micro {{Lie}} Theory for State Estimation in Robotics},
  author = {Sol{\`a}, Joan and Deray, Jeremie and Atchuthan, Dinesh},
  year = {2021},
  number = {arXiv:1812.01537},
  eprint = {1812.01537},
  primaryclass = {cs},
  publisher = {{arXiv}},
  urldate = {2022-10-25},
  archiveprefix = {arxiv},
  langid = {english}
}

@article{strabalaSeamlessHumanRobotHandovers2013,
  title = {Towards {{Seamless Human-Robot Handovers}}},
  author = {Strabala, Kyle Wayne and Lee, Min Kyung and Dragan, Anca Diana and Forlizzi, Jodi Lee and Srinivasa, Siddhartha and Cakmak, Maya and Micelli, Vincenzo},
  year = {2013},
  journal = {Journal of Human-Robot Interaction},
  volume = {2},
  number = {1},
  pages = {112--132},
  issn = {2163-0364},
  doi = {10.5898/jhri.2.1.strabala}
}

@inproceedings{vanduijkerenPathfollowingNMPCSeriallink2016,
  title = {Path-Following {{NMPC}} for Serial-Link Robot Manipulators Using a Path-Parametric System Reformulation},
  booktitle = {Proceedings of the {{European Control Conference}}},
  author = {Van Duijkeren, Niels and Verschueren, Robin and Pipeleers, Goele and Diehl, Moritz and Swevers, Jan},
  year = {2016},
  pages = {477--482},
  address = {{Aalborg, Denmark}},
  doi = {10.1109/ECC.2016.7810330},
  urldate = {2023-03-31},
  langid = {english}
}

@inproceedings{widmannHumanMotionPrediction2018,
  title = {Human {{Motion Prediction}} in {{Human-Robot Handovers}} Based on {{Dynamic Movement Primitives}}},
  booktitle = {Proceedings of the {{European Control Conference}} ({{ECC}})},
  author = {Widmann, Dominik and Karayiannidis, Yiannis},
  year = {2018},
  month = jun,
  pages = {2781--2787},
  publisher = {{IEEE}},
  address = {{Limassol}},
  doi = {10.23919/ECC.2018.8550170},
  urldate = {2022-11-23},
  isbn = {978-3-9524269-8-2},
  langid = {english}
}

@article{wuOnlineMotionPrediction2019,
  title = {On-Line {{Motion Prediction}} and {{Adaptive Control}} in {{Human-Robot Handover Tasks}}},
  author = {Wu, Min and Taetz, Bertram and Saraiva, Ernesto Dickel and Bleser, Gabriele and Liu, Steven},
  year = {2019},
  journal = {Proceedings of the IEEE Workshop on Advanced Robotics and its Social Impacts, ARSO},
  pages = {1--6},
  issn = {21627576},
  doi = {10.1109/ARSO46408.2019.8948750},
  isbn = {9781728131764}
}

@inproceedings{yangModelPredictiveControl2022,
  title = {Model {{Predictive Control}} for {{Fluid Human-to-Robot Handovers}}},
  booktitle = {Proceedings of the {{International Conference}} on {{Robotics}} and {{Automation}} ({{ICRA}})},
  author = {Yang, Wei and Sundaralingam, Balakumar and Paxton, Chris and Akinola, Iretiayo and Chao, Yu-Wei and Cakmak, Maya and Fox, Dieter},
  year = {2022},
  month = may,
  pages = {6956--6962},
  publisher = {{IEEE}},
  address = {{Philadelphia, PA, USA}},
  doi = {10.1109/ICRA46639.2022.9812109},
  urldate = {2022-09-08},
  isbn = {978-1-72819-681-7},
  langid = {english}
}

@article{zacharakiSafetyBoundsHuman2020,
  title = {Safety Bounds in Human Robot Interaction: {{A}} Survey},
  shorttitle = {Safety Bounds in Human Robot Interaction},
  author = {Zacharaki, Angeliki and Kostavelis, Ioannis and Gasteratos, Antonios and Dokas, Ioannis},
  year = {2020},
  month = jul,
  journal = {Safety Science},
  volume = {127},
  pages = {104667},
  issn = {09257535},
  doi = {10.1016/j.ssci.2020.104667},
  urldate = {2022-11-18},
  langid = {english}
}

@inproceedings{oelerichVDI,
  title = {Model Predictive Trajectory Planning for Human-Robot Handovers},
  booktitle = {Proceedings of VDI Mechatroniktagung},
  author = {Oelerich, Thies and Hartl-Nesic, Christian and Kugi, Andreas},
  year = {2024},
  pages = {65-72},
  langid = {english},
  url = {https://www.vdi-mechatroniktagung.rwth-aachen.de/global/show_document.asp?id=aaaaaaaacjcayqj&download=1},
  urldate = {2024-04-05},
}

\clearpage  % Verhindert automatischen Umbruch auf der letzten Seite. Bei Bedarf entfernen.

\end{document}